\documentclass[arxiv]{melba}

\usepackage{amsmath,amsfonts}

\usepackage{xcolor}
\usepackage[normalem]{ulem}
\usepackage{todonotes}
\usepackage{standalone}
\usepackage{placeins}
\usepackage{standalone}

\melbaid{YYYY:NNN}  
\doi{10.59275/j.melba.2024-AAAA}
\melbaauthors{Leclercq et al.}  
\email{a.corroyer-dulmont@baclesse.unicancer.fr}
\volume{3}
\firstpageno{1}  
\melbayear{2026}  
\datesubmitted{yyyy-m1-d1}  
\datepublished{yyyy-m2-d2}  

\melbaspecialissue{Medical Imaging with Deep Learning (MIDL) 2020}
\melbaspecialissueeditors{Marleen de Bruijne, Tal Arbel, Ismail Ben Ayed, Hervé Lombaert}

\ShortHeadings{CFB-GBM  v2.0}{LECLERCQ, et al.}

\title{CFB-GBM v2.0: An Augmented Longitudinal Dataset for Multi-Modal Glioblastoma Segmentation, Radiomics, and RANO Progression Tracking}

\author{
	\firstname Alexandre G. \surname LECLERCQ\aff{1,2,3}\orcid{0009-0004-6862-6613},
	\name Noémie N. Moreau\aff{1,2}\orcid{0009-0005-2117-2519}
	\name Hugo Audebert\aff{4}
    \name Andros Nassar\aff{4}
    \name Thomas Cochin\aff{2,4}
    \name Thomas Leleu\aff{2,4}
    \name Loïc Le Henaff\aff{2,5}
    \name Alexis Desmonts\aff{1,2,6}\orcid{0009-0009-4664-8770}
    \name Yoann Poirier\aff{5}
    \name Aurélie Dubru\aff{5}
    \name Laura Guillemette\aff{6}
    \name Pascal Lecoeur\aff{6}
    \name Kévin Lemasson\aff{6}
    \name Cyril Jaudet\aff{6}\orcid{0000-0003-2021-1983}
    \name Sébastien Bougleux\aff{3}\orcid{0000-0002-4581-7570}
    \name Romain Hérault\aff{3}\orcid{0009-0005-6369-8551}
    \name Carole Brunaud\aff{2}\orcid{0000-0002-5084-1479}
    \name Samuel Valable\aff{2}\orcid{0000-0003-0355-0270}
    \name Dinu Stefan\aff{4}\orcid{0000-0003-4537-2871}
    \name Charlotte Raboutet\aff{2,5}
    \name Alain Batalla\aff{6}\orcid{0000-0001-8489-5689}
    \name Joëlle Lacroix\aff{5}\orcid{0009-0005-7956-7732}
    \name Roman Rouzier\aff{7}\orcid{0000-0002-8167-6808}
    \name Aurélien Corroyer-Dulmont\aff{1,2,6}\orcid{0000-0002-9826-6565}
}
\affiliations{
	\num 1 \addr Artificial Intelligence Department, Centre François Baclesse, 14000 Caen, France \\
	\num 2 \addr Université de Caen Normandie, CNRS, Normandie Univ, ISTCT UMR6030, CYCERON, F-14000 Caen, France \\
	\num 3 \addr Université Caen Normandie, ENSICAEN, CNRS, Normandie Univ, GREYC UMR 6072, F-14000 Caen, France \\
	\num 4 \addr Radiation Oncology Department, Centre François Baclesse, 14000 Caen, France \\
	\num 5 \addr Radiology Department, Centre François Baclesse, 14000 Caen, France \\
    \num 6 \addr Medical Physics Department, Centre François Baclesse, 14000 Caen, France \\
    \num 7 \addr Surgery Department, Centre François Baclesse, 14000 Caen, France
}

\abstract{
    Glioblastoma (GBM) is the most aggressive primary brain tumor in adults, with a median overall survival of 15 months. Longitudinal, multi-modal imaging datasets with comprehensive clinical and treatment data are essential to support the development of reproducible computational methods for treatment response prediction, disease progression modelling, and personalized medicine. We present CFB-GBM v2.0, an extension of our previously released CFB-GBM dataset comprising 264 GBM patients treated according to the standard Stupp protocol.
    The primary contribution of this release is the completion of Gross Tumour Volume (GTV) delineations across all available timepoints ($t_0$, $t_1$ and $t_2$), increasing the overall GTV completion rate from 35\% to 97\%. This was achieved using a nnU-Net model pre-trained on BraTS 2021 and fine-tuned on CFB-GBM ground-truth contours, with the generated segmentations validated by five radiation oncologists. From these longitudinal GTV annotations, volumetric RANO 2.0 response category labels were derived for all available temporality pairs ($t_0 \rightarrow t_1$, $t_0 \rightarrow t_2$ and $t_1 \rightarrow t_2$). To further ease dataset usability and reproducibility, brain masks computed with HD-BET and pre-computed radiomic features extracted with PyRadiomics are provided for each patient timepoint and MRI modality. Additionally, the WHO classification guideline (2016 vs. 2021) applicable to each patient's diagnosis is now explicitly documented. CFB-GBM v2.0 is publicly available on The Cancer Imaging Archive (TCIA) at \url{www.cancerimagingarchive.net/collection/cfb-gbm}}


\begin{document}

\twocolumn[\maketitle]

\section{Introduction}
    Glioblastoma (GBM) is the most common and aggressive primary malignant brain tumor in adults. Characterized by cellular and genetic heterogeneity, the current standard treatment consists of surgical resection followed by the Stupp protocol \citep{stuppRadiotherapyConcomitantAdjuvant2005}, comprising concurrent radiotherapy and temozolomide (TMZ) chemotherapy, followed by six cycles of adjuvant TMZ. Despite this aggressive multimodal intervention, patient therapeutic response remains highly unpredictable, resulting in a critically poor prognosis with a median overall survival (OS) of approximately 15 months. This substantial inter-patient variability in treatment efficacy underscores a critical, unmet clinical need for personalized medicine paradigms capable of adapting to the unique evolutionary trajectory of each patient’s tumor.
    
    Magnetic Resonance Imaging (MRI) plays an indispensable role throughout the clinical pipeline, serving as the primary modality for initial diagnosis, non-invasive tumor characterization, and the longitudinal evaluation of treatment response. In current clinical practice, determining treatment efficacy requires a delayed observational period of up to two months, during which clinicians must differentiate treatment-induced inflammatory changes, such as pseudoprogression, from true disease progression. This delay carries particular clinical significance given the already limited median OS of GBM patients. A critical unmet need therefore remains: the early and reliable identification of treatment responders versus non-responders, along with improved characterization of intra- and inter-tumoral spatial heterogeneity, to enable timely therapeutic adjustments within a personalized medicine framework. While several studies have investigated this challenge using either private or publicly available datasets \citep{chenPredictionMGMTMethylation2025, moreauEarlyCharacterizationPrediction2025, moya-saezSyntheticMRIImproves2022}, the predominant reliance on proprietary data limits the comparability and reproducibility of results across the literature. Although publicly available datasets are essential for benchmarking and cross-study comparison, their availability for GBM remains limited \citep{cepedaRioHortegaUniversity2023a, suterLUMIEREDatasetLongitudinal2022, bakasUniversityPennsylvaniaGlioblastoma2022}.
    
    To address these limitations, we previously released CFB-GBM \citep{TCIA} 
    , a longitudinal, multi-modal dataset assembled from the clinical routine of 264 patients newly diagnosed with GBM and treated in a French oncology center according to the standard Stupp protocol between 2017 and 2023. 
    In addition to multi-modal MRI acquired at multiple timepoints ($t_0$, $t_1$ and $t_2$), the dataset incorporates planning CT scans, radiotherapy dose maps, and Gross Tumor Volume (GTV) delineations at the pre-treatment timepoint ($t_0$), alongside structured clinical data. This combination of multi-modal longitudinal imaging with comprehensive treatment and clinical metadata constitutes a distinctive feature relative to other publicly available GBM datasets such as LUMIERE  \citep{suterLUMIEREDatasetLongitudinal2022} and UPenn-GBM \citep{bakasUniversityPennsylvaniaGlioblastoma2022}.
    This characteristic is particularly relevant for tasks such as tumor control probability modelling \citep{yorkeModelingClinicalOutcomes2023} and automated radiotherapy plan generation \citep{goodingFullyAutomatedRadiotherapy2024}, in which the precise characterization of the delivered treatment is a central modelling factor. Moreover, the breadth and completeness of the available data support a wide range of research objectives, including treatment outcome prediction from pre-treatment information \citep{moreauEarlyCharacterizationPrediction2025}, cross-modality image synthesis (MRI to CT \citep{boulangerDeepLearningMethods2021}; intra-MRI modality translation \citep{kimAdaptiveLatentDiffusion2024}), and MRI resampling and denoising \citep{moummadImpactResamplingDenoising2022}.

    In the literature, personalized medicine is further approached through Disease Progression Modelling (DPM), a paradigm that aims to predict the temporal evolution of a disease from longitudinal patient data. A key instantiation of this paradigm is Treatment Response Prediction (TRP), which focuses on quantifying the impact of therapeutic interventions on disease trajectory. Various deep learning approaches leveraging multi-timepoint MRI data and tumor segmentation have been explored for this purpose \citep{xiaoIntracerebralHaemorrhageGrowth2021}.
    In standard clinical practice, treatment efficacy is assessed according to the Response Assessment in Neuro-Oncology (RANO) criteria \citep{wenRANO20Update2023}, which quantifies active tumor burden based on contrast-enhancing regions on T1-weighted MRI. RANO-based response classification has recently emerged as a distinct TRP subtask, as exemplified by the 2025 BraTS sub-challenge on Brain Tumor Progression (BraTSPRO), which proposed a classification task for RANO criteria prediction leveraging the LUMIERE dataset \citep{suterLUMIEREDatasetLongitudinal2022}.

\begin{figure*}[t]
		\centering
		\includegraphics[trim=1cm 24cm 1cm 3cm, clip, width=1\linewidth]{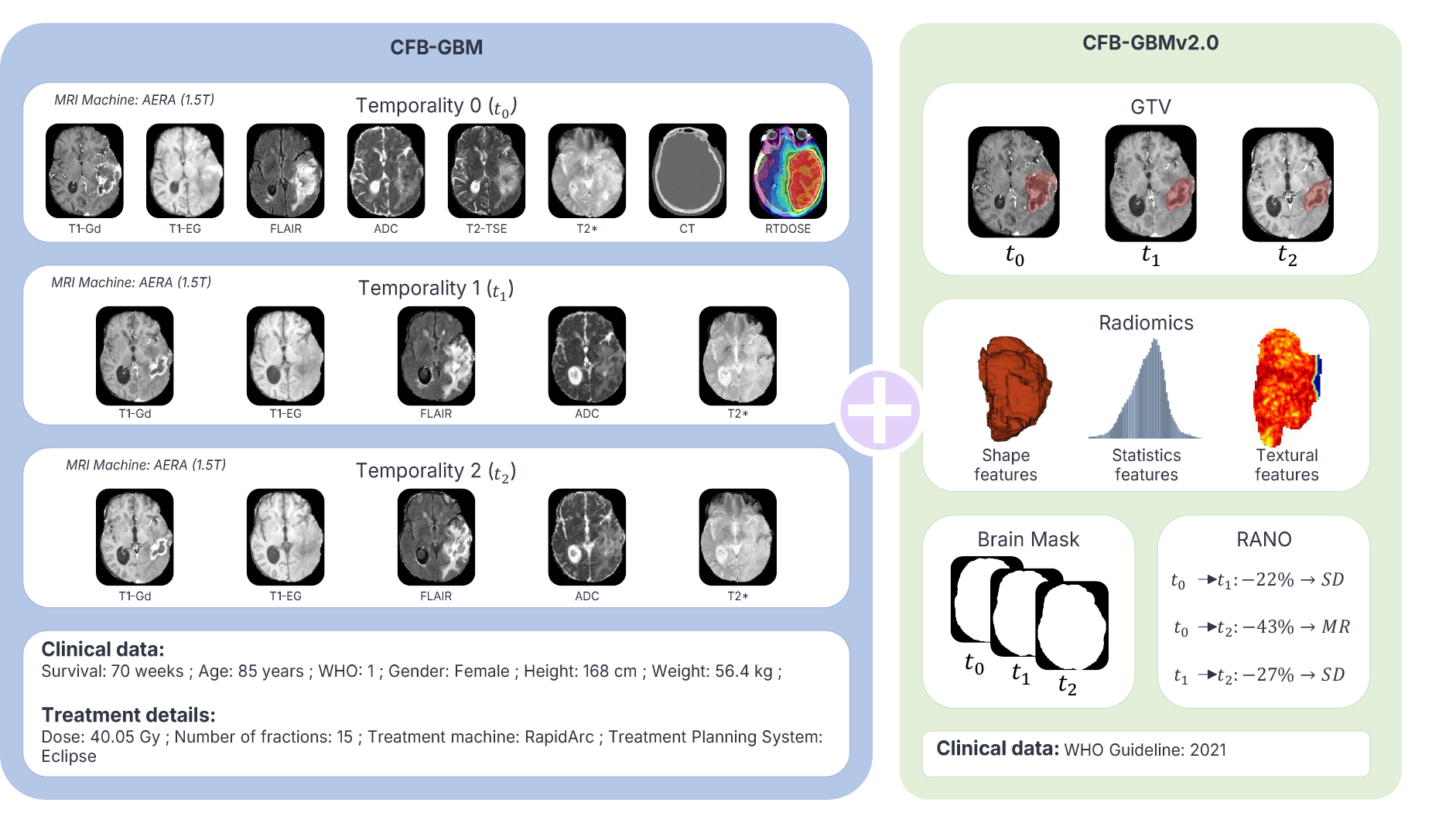}
		\caption{CFB-GBM v2.0: dataset extension overview}
		\label{graphical_abstract}
	\end{figure*}

    Motivated by these needs, we propose CFB-GBM v2.0, an extension of our initial release (\textbf{Figure~\ref{graphical_abstract}}). The primary contributions of this updated dataset are as follows:
    \begin{itemize}
      \item \textbf{Gross Tumor Volume (GTV)} segmentation at multiple timepoints, completing missing annotations from the original release and enabling longitudinal DPM tasks dependent on serial tumor segmentation.
      \item \textbf{RANO criteria annotations} derived from multi-timepoint tumor segmentations, directly supporting the RANO classification prediction task.
      \item \textbf{Pre-computed radiomic features} for each patient timepoint and MRI modality, providing ready-to-use quantitative descriptors widely employed across downstream tasks.
      \item \textbf{Brain masks} for each patient timepoint, facilitating intensity normalization and supporting AI-based MRI processing pipelines.
      \item \textbf{World Health Organization (WHO) classification information}, specifying whether each patient was diagnosed under the WHO 2016 or WHO 2021 guidelines, a clinically important distinction given the revised 2021 definition of GBM, which now requires IDH-wildtype status.
    \end{itemize}
    
\section{The Baseline Cohort: CFB-GBM}

\subsection{Patient Population \& Demographics}
    The CFB-GBM dataset included 264 patients newly diagnosed with GBM and treated with the standard Stupp protocol at our oncology center (Centre François Baclesse) between November 2017 and December 2023. Patients were recruited based on the following inclusion criteria: (i) age $\geqslant$ 18 years old, (ii) deceased patients, (iii) not enrolled in a clinical trial and (iv) imaging data acquired at our center. Clinical characteristics for each patient are summarized in \textbf{Table~\ref{patient_population}}.
    
    \begin{table}[ht]
        \centering
        \caption{Demographic data description}\label{patient_population}
        
    \begin{tabular}{crl}
        			\textbf{Demographics} & \multicolumn{2}{c}{\textbf{Values}}\\
        			\hline
        			\multirow{2}{*}{Gender} & Male: & n=$158$\\
        			& Female:& n=$106$\\
        			\hline
         			Age at the diagnosis & Min-Max: & $[25.0 - 90.0]$\\
        			(years) & Mean $\pm$ std:& $67.0 \pm 11.0$\\
        			\hline
         			\multirow{2}{*}{Height (cm)} & Min-Max: & $[146.0 - 191.0]$\\
        			& Mean $\pm$ std: & $168.9 \pm 9.3$\\
        			\hline
        			\multirow{2}{*}{Weight (kg)} & Min-Max: & $[37.7 - 116.0]$\\
        			& Mean $\pm$ std: & $72.3 \pm 13.8$\\
        			\hline
        			Overall survival & Min-Max: & $[3.0 - 238.0]$\\
        			(weeks)& Mean $\pm$ std: & $57.5 \pm 43.3$\\
        			\hline
        			\multirow{7}{*}{WHO Performance Status} & 0: & n=$45$\\
        			& 1:& n=$132$ \\
        			& 2:& n=$70$ \\
        			& 3:& n=$16$ \\
        			& 4:& n=$0$ \\
        			& 5:& n=$0$ \\
        			& Unknown:& n=$1$ \\
        			\hline
                    \multirow{2}{*}{WHO guideline} & 2016: & n=185\\
        			& 2021:& n=79\\
        			\hline
        		\end{tabular}

    \end{table}
    

\subsection{Imaging Modalities \& Acquisition}
    Depending on their availability, the imaging studies performed for each patient include: T1-Gadolinium (T1-Gd), T1 anatomical, T1 gradient echo (T1-GE), T2 anatomical, T2-FLAIR, T2*, apparent diffusion coefficient map (ADC map), CT scans, GTV and radiation dose distribution (RTDOSE).
    
\subsubsection{MRI Acquisition}
    MRI was performed on AREA/VIDA SIEMENS 1.5/3 Tesla magnets using a brain dedicated 16 channels coil with the patient in a supine position. Prior to the examination, patients were injected with 0.2 mL/kg of DOTAREM (500$\mu$mol/ml). A shimming process was made before acquisition of several sequences: 
    \begin{itemize}
        \item \textit{T1-Gd:} Tumor gadolinium enhancement was detected with a post-Gd T1 brain sequence using the following parameters: TR/TE$_{eff}$ = 2070/3.15 msec; angle = 15°; NEX = 1; acquisition time = 4:48 (min:ss); 208 contiguous slices; 3D resolution = 0.5$\times$0.5$\times$1 mm and acquisition matrix = 512$\times$512 pixels.
        
        \item \textit{T1:} Anatomical T1 contrast was obtained with a turbo spin echo sequence with the parameters listed below: TR/TE$_{eff}$ = 686/11 msec; angle = 142°; NEX = 2; acquisition time = 1:50 (min:ss); 30 contiguous slices; 3D resolution = 0.72$\times$0.72$\times$4 mm and acquisition matrix = 260$\times$320 pixels.
        
        \item \textit{T1-GE:} Sequence parameters: TR/TE$_{eff}$ = 313/4.76 msec; angle = 90°; 3D resolution = 0.65$\times$0.65$\times$4 mm; NEX = 1; acquisition time = 2:20 (min:ss); 30 contiguous slices and acquisition matrix = 300$\times$352 pixels.
        
        \item \textit{T2:} Sequence parameters: TR/TE$_{eff}$ = 7410/107 msec; angle = 166°; NEX = 1; acquisition time = 1:50 (min:ss); 30 contiguous slices; 3D resolution = 0.76$\times$0.76$\times$4 mm and acquisition matrix = 288$\times$320 pixels.
        
        \item \textit{T2-FLAIR:} Sequence parameters: TR/TE$_{eff}$ = 8150/134 msec; angle = 180°; NEX = 2; acquisition time = 1:55 (min:ss); 30 contiguous slices; 3D resolution = 1$\times$1$\times$4 mm and acquisition matrix = 184$\times$256 pixels.
        
        \item \textit{T2*:} Sequence parameters: TR/TE$_{eff}$ = 960/25 msec; angle = 20°; NEX = 1; acquisition time = 2:15 (min:ss); 30 contiguous slices; 3D resolution = 0.5$\times$0.5$\times$4 mm and acquisition matrix = 464$\times$512 pixels. 
        
        \item \textit{ADC map:} Sequence parameters: TR/TE$_{eff}$ = 2710/75.2 msec; angle = 180°; NEX = 1; acquisition time = 1:50 (min:ss); 30 contiguous slices; 3D resolution = 0.75$\times$0.75$\times$4 mm; acquisition matrix = 320$\times$320 pixels, B-value = 1000 and number of directions = 4.
      \end{itemize}

\subsubsection{CT Scans Acquisition}
    We acquired axial CT images using SIEMENS Confidence$\text{\textregistered}$ and PHILIPS Big Bore$\text{\textregistered}$ scanners, with the following parameters:
\begin{itemize}
    \item \textit{SIEMENS Confidence$\text{\textregistered}$:} Slice thickness = 2mm; tube voltage = 120kV; gantry rotation = 0.8 (s); acquisition time = 2:50 (min:ss); 208 contiguous slices; 3D resolution = 1$\times$1 mm and acquisition matrix = 512$\times$512 pixels.
    
    \item \textit{PHILIPS Big Bore$\text{\textregistered}$:} Slice thickness = 2mm; tube voltage = 120kV; acquisition time = 1:58 (min:ss); 188 contiguous slices; 3D resolution = 1.27$\times$1.27 mm and acquisition matrix = 512$\times$512 pixels.
\end{itemize}

\subsubsection{Treatment Planning Data Acquisition}
\subsubsection*{\textbf{Radiation Dose Distribution:}}
In accordance with the standard clinical practice, radiotherapy was delivered using Volumetric Modulated Arc Therapy (VMAT). The standard treatment consisted of 60 Gy delivered in 30 daily fractions (2 Gy/fraction) over six weeks (five days per week). For patients over 70 years old or with a WHO performance status $<$ 2, a hypofractionated regimen was administered, delivering a total dose of 40.05 Gy in 15 fractions (2.67 Gy/fraction) over three weeks. Radiation dose distribution were computed with Precision$\text{\textregistered}$-TPS, ECLIPSE$\text{\textregistered}$-TPS and Raystation$\text{\textregistered}$-TPS, and radiotherapy treatment was delivered using tomotherapy (Accuray$\text{\textregistered}$) and RapidArc (Varian$\text{\textregistered}$) machines. RTDOSE image details are as follows:  168 contiguous slices; 3D resolution = 2.5$\times$2.5$\times$5 mm and acquisition matrix = 167$\times$95 pixels. Treatment details are summarized in \textbf{Table~\ref{treatment_details}}.

\begin{table}[h]
        \centering
        \caption{Treatment details}\label{treatment_details}
        
    \begin{tabular}{crl}
        			\textbf{Radiation details} & \multicolumn{2}{c}{\textbf{Value}}\\
        			\hline
        			\multirow{4}{*}{Radiation dose} & 40.05 Gy: & n=$62$\\
        			& 60.00 Gy:& n=$123$\\
        			& Others:& n=$9$\\
        			& NA:& n=$70$\\
        			\hline
        			\multirow{4}{*}{Radiation fractions} & 15: & n=$63$\\
        			& 30:& n=$124$\\
        			& Others:& n=$7$\\
        			& NA:& n=$70$\\
        			\hline
        		\end{tabular}

    \end{table}

\subsubsection*{\textbf{Tumor Delineation:}} 
The GTV, representing the visible volume of the primary tumor, was manually delineated for each patient, in accordance with the European Society for Radiotherapy and Oncology Advisory Committee on Radiation Oncology Practice guidelines. 

\subsubsection{Data Preprocessing \& Standardization}
\subsubsection*{\textbf{Image Format and Conversion:}}
All images were retrieved from the Picture Archiving Communication System (PACS) in Digital Imaging and Communications in Medicine (DICOM) format and converted into the compressed Neuroimaging Informatics Technology Initiative (NifTI) file format (.nii.gz) to avoid leakage of sensitive metadata from the original DICOM headers. To preserve the integrity of the data and prevent processing bias, no intensity normalization or further image modifications were performed. 

\subsubsection*{\textbf{Image Registration:}}
To ensure spatial alignment, all imaging modalities were registered and resampled in the same space with rigid registration, using the baseline T1-Gd sequence as the reference. This rigid registration was implemented using the SimpleITK \citep{simpleITK1} library, applying a three-step multi-resolution approach with the following parameters: Mattes Mutual Information (with 50 bins) as the optimization metric, along with configured shrink factors and smoothing sigmas. 

\subsubsection*{\textbf{Skull Stripping \& Facial Defacing:}}
To ensure patient privacy and prevent identity reconstruction from skull morphology, we applied a brain extraction method, from ANTsPyNet \citep{antspy}, on each MRI sequence. For CT scans, a two-step anonymization was performed: (i) TotalSegmentator \citep{TotalSegmentator} was used to remove the radiation therapy thermoplastic masks, and (ii) CTA-DEFACE \citep{CTADEFACE} model was applied to the TotalSegmentator segmented images. 

\subsection{Longitudinal Timeline}

\begin{figure}[h]
        \centering
        \includegraphics[trim=5cm 33.5cm 6cm 6cm, clip, width=0.55\textwidth]{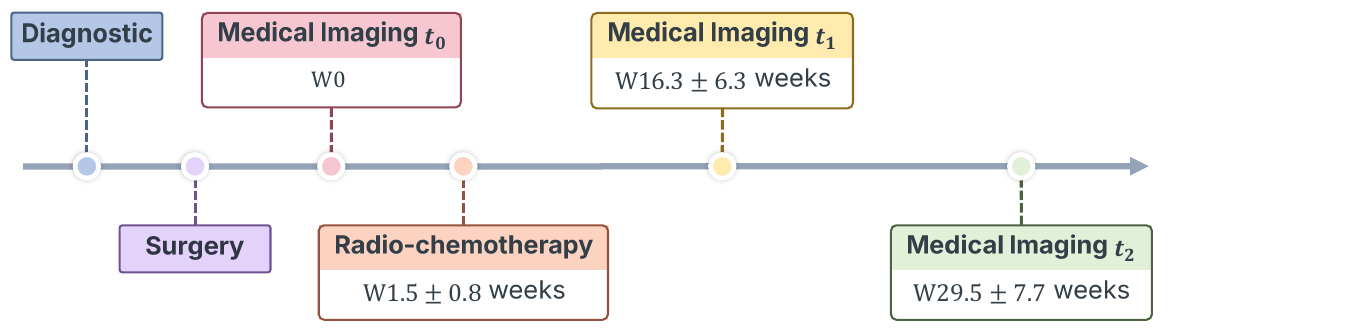}\\
        \caption{Timeline of longitudinal medical imaging acquisition}
        \label{timeline}
    \end{figure}

    A major asset of the CFB-GBM dataset is its inclusion of MRI scans obtained at different times: one week before radio-chemotherapy (corresponding to temporality 0 (t$_{0}$)), and approximately four (temporality 1 (t$_{1}$)) and six months (temporality 2 (t$_{2}$)) after (t$_{0}$) MRI acquisition (\textbf{Figure~\ref{timeline}}). \\

    The number of imaging modalities at different times is summarized in \textbf{Table~\ref{longitudinal}}.

    \begin{table}[h]
        \centering
        \caption{Overview of Available Imaging Modalities}\label{longitudinal}
        
    		\begin{tabular}{llll}
    			\textbf{Imaging modality} & \textbf{t$_{0}$} & \textbf{t$_{1}$} & \textbf{t$_{2}$} \\
    			\hline
    			T1-Gd & 264 & 163 & 118 \\
    			T1-GE & 215 & 126 & 82 \\
    			T1 & 0 & 17 & 14 \\
    			T2-FLAIR & 255 & 167 & 114 \\
    			T2 & 125 & 15 & 11 \\
    			T2* & 241 & 136 & 94 \\
    			ADC & 129 & 132 & 91 \\
    			CT & 195 & - & - \\
    			RTDOSE & 194 & - & - \\
    			GTV & 191 &- & - \\
    			\hline
    		\end{tabular}

    \end{table}

All data files are structured according to the following naming convention:
\begin{itemize}
    \item \textbf{Folder:} CFB-GBM/patient/temporality/file
    \item \textbf{File format:} \textit{patient$\_$temporality$\_$modality.nii.gz}
\end{itemize}

\section{Methodology \& Validation}
	\subsection{Tumor Segmentation}
	    To complete the missing longitudinal GTV delineations in the original dataset, we trained an automatic segmentation model based on the nnU-Net framework \citep{isenseeNnUNetSelfconfiguringMethod2021}, which serves as a strong and well-established baseline for medical image segmentation tasks.
	    Standard GBM segmentation models trained on the BraTS 2021 dataset \citep{baidRSNAASNRMICCAIBraTS20212021} rely on four MRI modalities: T1, T1-CE, T2, and T2-FLAIR. However, MRI modality availability in CFB-GBM is inconsistent across patients and timepoints, making full four-modality inference impractical. We therefore trained a segmentation model using only T1-Gd and T2-FLAIR sequences, which constitute the most consistently available modality pair across the cohort.
	    The BraTS segmentation protocol decomposes the tumor into three sub-regions: the Gd-enhancing tumor (ET), the necrotic tumor core (NCR), and the peritumoral oedematous/invaded tissue (ED). To align with the clinical definition of GTV, we mapped the ET and NCR labels to a single GTV label, as both sub-regions correspond to tumor components visible on contrast-enhanced T1-weighted MRI. The ED region, which typically requires additional modalities for reliable delineation, was excluded from the GTV definition.
	    Following the standard nnU-Net V2 training pipeline, we evaluated three model configurations: a model trained exclusively on BraTS 2021, a model trained exclusively on the CFB-GBM GT segmentations, and a model pre-trained on BraTS 2021 and fine-tuned on the CFB-GBM ground-truth (GT) segmentations. Models were evaluated on the CFB-GBM test set using the Dice Similarity Coefficient (DSC) between predicted and GT segmentations. As reported in \textbf{Table~\ref{evaluation_model}}, the model pre-trained on BraTS and fine-tuned on our dataset training set achieves the best performance.

	    \begin{table}[h]
            \centering
        	\caption{Evaluation of Segmentation Models}\label{evaluation_model}
            
    		\begin{tabular}{lc}
    			\textbf{Training Dataset} & \textbf{Dice Score} \\
    			\hline
    			BRATS 2021 & $0.6128$ \\
    			CFB-GBM & $0.7860$ \\
    			\textbf{BRATS 2021 + CFB-GBM} & $\textbf{0.8031}$ \\
    			\hline
    		\end{tabular}

        \end{table}
        To further assess the quality of the automatically generated GTV delineations with the best model, we conducted an expert validation study on a subset of cases for which no ground truth was available (i.e., cases not seen during training). Specifically, 44 of the generated GTVs were sampled, with cases selected to ensure equal distribution across the three available imaging timepoints. Five radiation oncologists were asked to review and correct each sampled segmentation. 
        Three complementary metrics were computed between each model-generated GTV and the corresponding expert-corrected contour to assess segmentation quality.
	    
	    
	    \subsubsection*{\textbf{Dice Similarity Coefficient (DSC):}}
	    The DSC measures the global volumetric overlap between the model prediction and the expert-corrected segmentation, and is defined as:
	    $$DSC=\frac{2|GTV_{gen} \cap GTV_{corr}|}{|GTV_{gen}| + |GTV_{corr}|}$$
	    A DSC of 1.00 indicates a 100\% overlap between the two considered volumes.

	    \subsubsection*{\textbf{95th Percentile Hausdorff Distance (HD95)}}
	    To evaluate the local boundary distances and catch worst-case delineation errors without being overly sensitive to sporadic single-voxel outliers, we use the 95th percentile of the Hausdorff Distance (HD95). 
	    Let $X$ and $Y$ represent the surface point sets of $GTV_{gen}$ and $GTV_{corr}$, respectively. The directed Hausdorff distance from $X$ to $Y$ is defined as:
	    $$d_H(X, Y) = \max_{x \in X} \min_{y \in Y} \|x - y\|_2$$
	    The bidirectional HD95 is then computed as the 95th percentile of the combined distances from both directions:
	    $$HD95 = P_{95} \left( \{ \min_{y \in Y} \|x - y\|_2 \}_{x \in X} \cup \{ \min_{x \in X} \|y - x\|_2 \}_{y \in Y} \right)$$
	    A lower HD95 indicates closer boundary agreement, with 0 mm corresponding to a perfect match.
	    
	    \subsubsection*{\textbf{Average Symmetric Surface Distance (ASSD)}}
	    Complementing HD95, which captures localised worst-case deviations, the Average Symmetric Surface Distance (ASSD) quantifies the mean boundary error across the entire tumor surface. Using the same point sets $X$ and $Y$, it is defined as:
	
	    \begin{align*}
	        d(x, Y) &= \min_{y \in Y} \|x - y\|_2 \\
	        ASSD(X, Y) &= \frac{\sum_{x \in X} d(x, Y)  + \sum_{y \in Y} d(X, y)}{|X| + |Y|}
	    \end{align*}
	    The ASSD provides a comprehensive overview of the overall tightness of the fit, where a value closer to $0\text{ mm}$ indicates superior boundary agreement across the entire surface.
	    
    	\begin{table}[h]
            \centering
        	\caption{Radiation oncologist validation}\label{validation}
            
    		\begin{tabular}{lcccr}
    			\textbf{Expert} & \textbf{DSC ($\uparrow$)} & \textbf{HD95 ($\downarrow$)} & \textbf{ASSD ($\downarrow$)} & \textbf{Sample} \\
    			\hline
    			1 & $0.95\pm0.05$ & $3.87\pm2.92$ mm & $1.00\pm1.63$ mm & 7 \\
    			2 & $0.98\pm0.01$ & $1.76\pm0.68$ mm & $0.14\pm0.08$ mm & 11 \\
    			3 & $0.99\pm0.01$ & $0.82\pm0.62$ mm & $0.08\pm0.07$ mm& 5 \\
    			4 & $0.97\pm0.04$ & $1.69\pm1.36$ mm & $0.28\pm0.30$ mm&  16\\
    			5 & $0.99\pm0.01$ & $0.94\pm0.12$ mm & $0.07\pm0.04$ mm&  5\\
    			\hline
    			Overall & $0.97\pm0.03$ & $1.87\pm1.75$ mm & $0.31\pm0.74$ mm & 44
    		\end{tabular}

        \end{table}
        
        \textbf{Table~\ref{validation}} demonstrates strong agreement between the expert-corrected and model-generated GTV delineations, suggesting that the automatic segmentation model produces contours of sufficient quality for inclusion in the dataset.
        The gap between the expert-validation agreement (DSC of $0.97$, \textbf{Table~\ref{validation}}) and the test-set performance (DSC of $0.8031$, \textbf{Table~\ref{evaluation_model}}) stems mainly from the differing nature of the two reference contours. The manual ground-truth GTVs were delineated for radiotherapy planning and resampled into the common space, yielding staircase-like boundaries, whereas the model produces smooth contours. Scoring a smooth prediction against such a reference introduces a thin shell of boundary disagreement that Dice penalizes even at equal volume, an effect amplified for smaller lesions. In the validation, experts only lightly corrected these already-smooth contours, hence the higher overlap. This correction-based protocol, adopted because the five oncologists could not delineate all cases de novo, may introduce some anchoring, but this is unlikely to explain the full gap, which we attribute primarily to the boundary-character mismatch.
        
        The validated model was subsequently applied to generate the missing GTV delineations at $t_0$, $t_1$ and $t_2$. \textbf{Table~\ref{gtv}} summarizes the number of GTV annotations available in CFB-GBM v2.0 across all timepoints.
        
        
        \begin{table}[h]
            \centering
        	\caption{Imaging modalites with new additions}\label{gtv}
            
    	    \begin{tabular}{llll}
        		\hline
        	    GTV & $191{\rightarrow261}$ & ${0\rightarrow160}$ & ${0\rightarrow112}$ \\
        	    \hline
    		\end{tabular}

        \end{table}
		
	\subsection{RANO Criteria}
	RANO response categories were derived in accordance with the volumetric (3D) criteria defined in the RANO 2.0 guidelines \citep{wenRANO20Update2023}, using the contrast-enhancing tumor volume (the GTV, delineated on T1-Gd) as the reference measurable disease.
	For each patient, the relative volume change was computed for all available temporality pairs ($t_0 \rightarrow t_1$, $t_1 \rightarrow t_2$ and $t_0 \rightarrow t_2$), and is defined as:

	$$\text{Reduction Rate}_{t_i \rightarrow t_j} = 1-\frac{GTV_{t_j}}{GTV_{t_i}}, \quad \forall j > i$$

    A positive value indicates tumor shrinkage, while a negative value indicates growth. Each temporality pair was then assigned a RANO response category according to the following thresholds \citep{wenRANO20Update2023}:
	\begin{itemize}
	    \item \textbf{Complete Response (CR)}: No measurable enhancing lesion on target volume;
	    \item \textbf{Partial Response (PR)}: Volume reduction $\geq 65\%$;
	    \item \textbf{Stable Disease (SD)}: Volume reduction $< 65\%$ and volume increase $< 40\%$ (neither meeting PR nor PD criteria);
	    \item \textbf{Progressive Disease (PD)}: Volume increase $\geq 40\%$.
	\end{itemize}
    
	\subsection{Brain Mask}
	Brain masks were generated for each patient at each available timepoint using the HD-BET brain extraction tool \citep{isenseeAutomatedBrainExtraction2019} on T1-Gd MRI. 

	\subsection{Radiomics Features Extraction}
    For each MRI modality, radiomics features were extracted using the open-source PyRadiomics package \citep{PyRadiomics} (version 3.1.0), provided that the GTV was available. Prior to extraction, image preprocessing was performed, including intensity z-score normalization based on a brain mask (with the exception of ADC maps), scaling by a factor of 100 and a voxel resampling to an isotropic resolution of  $1\times1\times1$ mm$^{3}$. For a total of 2,424 MRIs, 107 original features were extracted in accordance with the Imaging Biomarker Standardization Initiative (IBSI) guidelines \citep{IBSI}. These features include: 18 first-order statistics features, 14 shape-based features and 75 texture features derived from gray-level matrices (24 Gray Level Co-occurrence Matrix features (GLCM); 16 Gray Level Run Length Matrix features (GLRLM); 16 Gray Level Size Zone Matrix features (GLSZM); Neighbouring Gray Tone Difference Matrix features (NGTDM) and Gray Level Dependence Matrix features (GLDM)). For gray-level discretization, a fixed bin width of 25 was applied to ADC maps, and a bin width of 5 was used for all other modalities \citep{suterLUMIEREDatasetLongitudinal2022}.

    

\section{Discussion \& Conclusion}
CFB-GBM v2.0 extends the original dataset with four key contributions designed to broaden its applicability to a wider range of computational oncology tasks. 

The completion of GTV delineations across all timepoints constitutes the most clinically significant addition, as serial tumor annotations are a prerequisite for Disease Progression Modelling and Treatment Response Prediction tasks. The automatic segmentation pipeline, based on a BraTS 2021 pre-trained nnU-Net fine-tuned on CFB-GBM ground-truth contours, demonstrated strong agreement with expert radiation oncologist corrections, supporting the reliability of the generated annotations. 

The provision of RANO 2.0 volumetric response categories derived from longitudinal GTV measurements directly addresses the growing interest in automated treatment response classification, as evidenced by recent benchmarking efforts such as the BraTSPRO 2025 challenge. 

Pre-computed radiomic features and brain masks further lower the barrier to entry for researchers working on quantitative imaging tasks, reducing preprocessing overhead and promoting reproducibility across studies leveraging the dataset.

Despite these contributions, CFB-GBM v2.0 retains a notable limitation inherited from the original cohort: the absence of molecular biomarker information, most critically IDH mutation status and MGMT promoter methylation. Both markers carry strong prognostic and predictive value in GBM, and their absence limits the dataset's utility for studies investigating genotype-phenotype relationships or stratifying patients by molecular subtype. This gap is further compounded by the change in the WHO classification of central nervous system tumors between the 2016 and 2021 editions, under which GBM is now restricted to IDH-wildtype tumors. 

Looking ahead, several extensions are envisioned for future iterations of the dataset. A natural next step would be the adoption of a BraTS-style multi-label tumor segmentation convention, decomposing the GTV into clinically and biologically distinct sub-regions (ET, NCR and ED label), which would enable a broader range of segmentation benchmarks and provide richer input features for downstream models. The prospective collection and integration of IDH mutation status and MGMT promoter methylation data would substantially increase the dataset's value for molecular outcome prediction tasks. These extensions would position CFB-GBM as a more comprehensive resource for the full spectrum of computational GBM research, from segmentation and radiomics to personalized treatment planning and survival prediction.







\acks{This study was funded by the Région Normandie through the “Booster IA” grant, NN. M and AG. L were supported by the Région Normandie.}

%

\ethics{This retrospective study received approval from the Centre François Baclesse institutional review board (internal ethics committee "CSE" data department, health data hub study number: F20240111162856). The study was conducted in compliance with the principles of the Declaration of Helsinki and the MR-004 guidelines established by the French National Institute for Health Data (INDS) for health research. Informed consent was obtained from all patients for the use of their data.}

\coi{We declare we don't have conflicts of interest.}

\data{CFB-GBM v2.0 is publicly available on The Cancer Imaging Archive (TCIA) at \url{https://doi.org/10.7937/v9pn-2f72}. It is a direct update of the original CFB-GBM dataset, and the TCIA collection also hosts the associated publication. 
Scripts used to produce the data are available at \url{https://github.com/AurelienCD/CFB-GBM}. The segmentation model used to generate the tumor delineations is publicly available on Hugging Face at \url{https://huggingface.co/AlexLECLERCQ/SegCFB-GBM}.}

\section{Licence}
CFB-GBM is licensed under \href{https://creativecommons.org/licenses/by/4.0/}{Creative Commons Attribution 4.0 International (CC BY 4.0)}.

\bibliography{sample}





\end{document}